%% file: main.tex
\documentclass[10pt,twocolumn,letterpaper]{article}
\usepackage{times}
\usepackage[letterpaper,textwidth=6.875in,textheight=8.875in,columnsep=0.3125in]{geometry}
\usepackage[dvipsnames]{xcolor}
\usepackage[numbers,sort&compress]{natbib}
\usepackage[format=plain,labelsep=period,font=small]{caption}
\usepackage[hyphens]{url}
\input{preamble}
\definecolor{linkblue}{rgb}{0.21,0.49,0.74}
\usepackage[breaklinks,colorlinks,allcolors=linkblue]{hyperref}

\title{\textbf{TwoPass: Draft-Guided Few-Step Video Prediction\\for Robotic Manipulation}}
\hypersetup{pdftitle={TwoPass: Draft-Guided Few-Step Video Prediction for Robotic Manipulation},pdfauthor={Yijun Liu, Shijia Ge, Tongtong Feng, Xin Wang, Yuan Meng, Yuwei Liu, Zhi Wang, Wenwu Zhu}}
\author{Yijun Liu \qquad Shijia Ge \qquad Tongtong Feng \qquad Xin Wang\\
Yuan Meng \qquad Yuwei Liu \qquad Zhi Wang\textsuperscript{\dag} \qquad Wenwu Zhu\textsuperscript{\dag}\\[4pt]
Tsinghua University\\[2pt]
{\normalsize \textsuperscript{\dag}Corresponding authors}}
\date{}
\makeatletter\providecommand{\@LN}[2]{}\providecommand{\@LN@col}[1]{}\makeatother
\begin{document}
\maketitle
\input{sections/0_abstract}
\input{sections/1_intro}
\input{sections/2_related}
\input{sections/3_method}

\input{sections/4_experiments}
\input{sections/5_conclusion}
{\small
\bibliographystyle{ieeenat_fullname}
\bibliography{references}
}
\clearpage
\appendix
\input{sections/6_appendix}

\end{document}

%% file: preamble.tex
\usepackage[T1]{fontenc}
\usepackage{graphicx}
\usepackage{amsmath,amssymb}
\usepackage{booktabs,multirow}
\usepackage{tikz}
\usetikzlibrary{arrows.meta,positioning,fit,backgrounds}
\usepackage{pgfplots}
\pgfplotsset{compat=1.18}
\usepackage{algorithm}
\usepackage{algpseudocode}
\newcommand{\method}{\textnormal{\scshape TwoPass}}
\newcommand{\nfe}[1]{\mbox{NFE\,=\,#1}}
\newcommand{\sg}{\operatorname{sg}}
\newcommand{\E}{\mathbb{E}}
\definecolor{draftblue}{HTML}{0072B2}
\definecolor{refineorange}{HTML}{D55E00}
\definecolor{feedbackgreen}{HTML}{009E73}

%% file: sections/0_abstract.tex
\begin{abstract}
Action-conditioned video prediction lets robots anticipate the visual consequences of their actions for planning and policy evaluation. Existing predictors denoise each future from noise without using their own estimate of it, which limits the quality of the predicted future, in particular of the manipulated objects, and requires many network calls. We present \method, a self-conditioned training-and-inference framework that keeps the standard flow-matching objective. Self-Generated Draft Training supplies a detached clean-video estimate from the shared network as conditioning at the same noisy state. A Refinement-Only Late Adapter applies a dedicated residual transformation when a draft is present. At inference, Draft-Reuse Iterative Refinement obtains both a solver update and the next draft from each velocity prediction, requiring as few as 4 function evaluations (NFE) per video chunk. Training samples discrete solver-query noise levels and updates the pretrained backbone. Extensive experiments on Bridge and Language Table show that \method\ achieves state-of-the-art results with 4 NFE, far fewer than the 50--59 calls of existing predictors. Code: \mbox{\url{https://github.com/PlantPotatoOnMoon/TwoPass}}.
\end{abstract}

%% file: sections/1_intro.tex
\section{Introduction}
\label{sec:intro}
\input{figures/motivation}
Action-conditioned video generation predicts future observations from a current scene and candidate robot controls, which supports planning, policy evaluation and data generation~\cite{ha2018worldmodels,irasim,unisim,dreamdojo}. Interest in such models has grown with embodied intelligence, where foundation and generative models increasingly support robot control and planning~\cite{du2023learninguniversalpoliciestextguided,liu2025spatialpolicyguidingvisuomotor,ge2026vggtdpgeneralizablerobotcontrol,ge2026makecodepolicygreat}, and with rapid advances in video generation~\cite{blattmann2023stablevideodiffusionscaling,yang2025cogvideoxtexttovideodiffusionmodels,wan}.

Planning and policy evaluation rely on such predictions~\cite{unisim,ctrlworld,dreamdojo}, and the decisions they produce depend on the quality of the predicted future where the scene changes: the manipulated objects and the robot arm (Fig.~\ref{fig:motivation}b). Current action-conditioned video generation methods~\cite{irasim,fu2026learningvideogenerationrobotic,yang2026orv4doccupancycentricrobot} still make errors in these regions, where manipulated objects can be misplaced, deformed or lost (Fig.~\ref{fig:external_candidate}). They inherit the iterative denoising of diffusion and flow-matching video generators~\cite{ho2022videodiffusionmodels,he2023latentvideodiffusionmodels,flowmatching,wan}: every step predicts the future again from a noisy state, over tens of network calls, without using the estimate of the future that the model has already produced.

Action-conditioned prediction of manipulation scenes raises three challenges. First, manipulated objects must keep their shape and appearance while they are moved, rotated and occluded by the robot arm; yet they occupy a small part of the frame and change quickly, and an error made on them early in generation tends to persist. Second, the details of these small objects must be preserved without sacrificing the overall scene and motion that the pretrained video model already captures; one predictor has to achieve both. Third, planning and policy evaluation predict repeatedly for many candidate actions, so these improvements must not come from additional network calls.

To address these challenges, we introduce \method, a self-conditioned training-and-inference framework (Fig.~\ref{fig:motivation}c). It builds on self-conditioning~\cite{analogbits,jabri2023scalableadaptivecomputationiterative,strudel2022selfconditionedembeddingdiffusiontext}, which feeds a model's own provisional estimate back as an input. For the first challenge, \textbf{Self-Generated Draft Training} constructs a draft, a detached clean estimate of the future from the current shared weights, and supplies it as an input to a supervised prediction at the same noisy state, with the standard flow velocity as target. The model thus learns to predict on top of its own estimate of the future, as at inference, under supervision from the actual future, so that manipulated objects stay consistent. For the second, a \textbf{Refinement-Only Late Adapter} adds a residual transformation before the shared prediction head, active only when a draft is present. It gives the correction of details its own pathway and leaves the original prediction path unchanged. For the third, \textbf{Draft-Reuse Iterative Refinement} passes each clean estimate to the next solver query, so one velocity prediction supplies both the state update and the next draft. Inference thus needs no extra network call for the draft, so the NFE equals the number of solver steps and repeated prediction stays cheap (Fig.~\ref{fig:motivation}a). We train on the noise levels that 4- and 6-NFE sampling visits and evaluate on the complete BridgeData V2 (Bridge)~\cite{bridge} and Language Table~\cite{language} test sets, against existing methods and matched ordinary-Wan~\cite{wan} and self-conditioning controls.

Our contributions are:
\begin{itemize}
\item A self-conditioned training scheme for action-conditioned video prediction, combining same-state detached drafts with a refinement-specific pathway under the standard flow-matching objective.
\item An inference procedure that reuses drafts in latent video flow prediction, coupling each solver update to the next conditioning draft, with role-dependent adapter activation and one network call per solver query.
\item Extensive experiments on Bridge and Language Table, where \method\ achieves state-of-the-art results with 4 NFE, far fewer than the 50--59 calls of existing predictors, together with analyses of training pass depth, NFE and individual components.
\end{itemize}

%% file: figures/motivation.tex
\begin{figure*}[t]
\centering
\includegraphics[width=\textwidth]{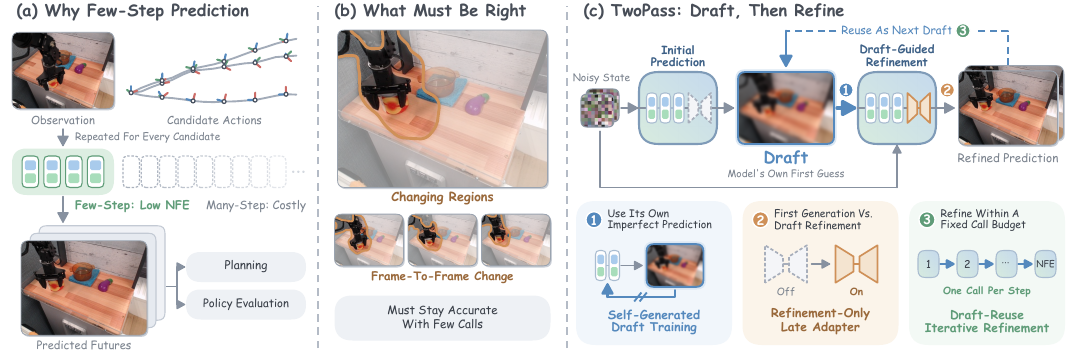}
\caption{\textbf{Motivation and concept of \method.} (a) Planning and policy evaluation repeatedly predict the outcomes of candidate actions, so prediction must be accurate and fast. (b) Prediction must capture changing regions and frame-to-frame visual changes; orange contours show simplified outer boundaries of the main regions that change between frames. (c) \textbf{Self-Generated Draft Training} teaches the model to use its own imperfect predictions; a \textbf{Refinement-Only Late Adapter} gives draft-conditioned correction its own pathway; and \textbf{Draft-Reuse Iterative Refinement} recycles clean estimates without extra network calls. The draft and supervised training passes share a noisy state, whereas inference advances the solver state between calls.}
\label{fig:motivation}
\end{figure*}

%% file: sections/2_related.tex
\section{Related Work}
\label{sec:related}
\paragraph{Video prediction and world models for manipulation.}
World models predict future observations in response to actions~\cite{ha2018worldmodels,bruce2024geniegenerativeinteractiveenvironments}, and many build on diffusion video generators~\cite{ho2022videodiffusionmodels,he2023latentvideodiffusionmodels,blattmann2023stablevideodiffusionscaling,yang2025cogvideoxtexttovideodiffusionmodels}. For manipulation, generated video supports planning and policy learning~\cite{du2023learninguniversalpoliciestextguided}. IRASim~\cite{irasim} and UniSim~\cite{unisim} predict visual action consequences; Ctrl-World combines frame-level action conditioning with pose-conditioned memory~\cite{ctrlworld}. RoboMaster~\cite{fu2026learningvideogenerationrobotic} guides manipulation videos with object masks and phased 2D robot--object trajectories; ORV~\cite{yang2026orv4doccupancycentricrobot} adds 4D semantic occupancy guidance. Large robot datasets~\cite{embodimentcollaboration2025openxembodimentroboticlearning} support the training of such models. We adapt the Wan video prior~\cite{wan,dit} with aligned actions and a pinned observation boundary, using Bridge and Language Table~\cite{bridge,language}.

\paragraph{Few-step video generation and sampling.}
Few-step generation relies on fast samplers~\cite{song2022denoisingdiffusionimplicitmodels,lu2025dpmsolverfastsolverguided,unipc} or on distilling a multi-step teacher~\cite{song2023consistencymodels,yin2024onestepdiffusiondistributionmatching,yin2025slowbidirectionalfastautoregressive}, and several robot world models generate with few steps~\cite{park2026overthinkingonestepgeneration,wang2026interactiveworldsimulatorrobot,dreamdojo,dreamxteam2026dreamxphi10actionconditionedvideo,jeon2026roboworldfastreliableneural}. Distillation first trains a multi-step teacher and then runs it, often with an auxiliary score network~\cite{yin2024onestepdiffusiondistributionmatching}, in a further training stage. \method\ instead trains one network with the standard flow-matching loss~\cite{flowmatching}, at a fixed noisy state and on the noise levels that its UniPC sampler visits~\cite{esser2024scalingrectifiedflowtransformers,curriculum}; it needs no teacher or auxiliary network, which avoids their training cost and keeps the memory footprint of a single network.

\paragraph{Self-conditioning and iterative refinement.}
Analog Bits conditions a denoising model on its clean-data estimate, generates training feedback without backpropagating through the draft, and recycles predictions at sampling time~\cite{analogbits}. Self-conditioning has since been applied to latent tokens in image and video diffusion~\cite{jabri2023scalableadaptivecomputationiterative} and to text diffusion~\cite{strudel2022selfconditionedembeddingdiffusiontext}. These operations are the foundation of our method, which instantiates them for action-conditioned latent video flow prediction on a pretrained backbone. \method\ differs in three respects. The draft enters through a separate zero-initialized patch embedding, so the pretrained function is preserved at initialization. A late adapter is active only in the refinement role, so the draft-conditioned prediction passes through a transformation that the draft-generating pass does not use. Training noise is drawn from the solver schedules of the few-step budgets at which the model is run. Plain self-conditioning with the same backbone, and self-conditioning with the adapter active in both roles, are therefore the central comparisons (Sec.~\ref{sec:wan_ablation}).

%% file: sections/3_method.tex
\section{Method}
\label{sec:method}
\input{figures/method}
\method\ is a training-and-inference framework in which an action-conditioned video predictor learns to use its own provisional future under the standard flow-matching objective (Fig.~\ref{fig:method}). On feedback-enabled updates, a shared-weight, no-gradient pass produces a clean draft; the supervised pass receives that draft at the same noisy state. A late adapter is active in the refinement role. During sampling, each solver query supplies both a velocity update and the next draft, preserving a budget of 4 or 6 NFE per chunk. The framework starts from a pretrained backbone and uses no additional distillation teacher or distribution-matching loss. Training samples discrete solver-query noise levels.

\subsection{Action-conditioned latent flow prediction}
Let $I_0$ be an observed image and $a_{1:T}$ the subsequent actions. A frozen video autoencoder encodes the target clip into $x_0\in\mathbb{R}^{C\times F\times H'\times W'}$. The subscript 0 denotes clean data; $x_0^{(0)}$ denotes the first latent frame. We use 17 image frames, corresponding to five latent frames. The observed image is independently encoded into the boundary latent $b$. The model predicts a velocity field conditioned on actions and the frozen text-encoder context $c$; $c$ encodes task text on Bridge and the empty prompt on Language Table.

The action sequence is aligned with the autoencoder's temporal groups. The first latent group receives the zero initial action, and group $g\geq1$ receives
\begin{equation}
\bar a_g=\frac{1}{4}\sum_{j=4g-3}^{4g}a_j.
\label{eq:actions}
\end{equation}
An action MLP projects each group to the six modulation vectors used by the transformer block, broadcasting them across that frame's spatial tokens. Its final projection is zero-initialized. All compared Wan variants use this same action representation.

For future latents, sample Gaussian noise $\epsilon$ and a noise/timestep pair $(\sigma,\tau)\sim q_{\rm grid}$ from the discrete query distribution (training-noise paragraph of Sec.~\ref{sec:eval}), then form the linear path
\begin{equation}
 x_\sigma=(1-\sigma)x_0+\sigma\epsilon,
 \qquad v^\star=\epsilon-x_0.
\label{eq:flow}
\end{equation}
We overwrite $x_\sigma^{(0)}$ with $b$ and exclude the observed latent frame from the training loss. Denote the future-frame mask by $M$. For a conditioning draft $d$ and role flag $r$, the model is $v_{\theta,\phi}(x_\sigma,\tau,a,c,d;r)$, where $\tau$ is the model timestep and $\phi$ are adapter parameters. The ordinary Wan baseline has neither a draft-input pathway nor an adapter.

\subsection{Self-Generated Draft Training}
\label{sec:feedback}
A Bernoulli variable $z\sim\operatorname{Bernoulli}(p)$ chooses whether an update uses a draft; $p=0.5$. If $z=1$, the current shared model first predicts a velocity without feedback or the refinement adapter:
\begin{align}
 \tilde v&=v_{\theta,\phi}(x_\sigma,\tau,a,c,0;0),\\
 d&=\sg\!\left(x_\sigma-\sigma\tilde v\right),
 \qquad d^{(0)}\leftarrow b.
\label{eq:draft}
\end{align}
The operator $\sg$ stops gradients, and the draft forward pass runs without an autograd graph. If $z=0$, the draft is zero and no draft pass is executed. The supervised prediction is
\begin{equation}
\begin{aligned}
 \hat v&=v_{\theta,\phi}(x_\sigma,\tau,a,c,d;z),\\
 \mathcal L&=\E\!\left[\frac{\|M\odot(\hat v-v^\star)\|_2^2}{\|M\|_1}\right],
\end{aligned}
\label{eq:loss}
\end{equation}
where the mask is broadcast over latent channels and spatial positions when counting its entries. The network learns from the velocity target, with the draft available as an additional input. The draft itself is neither a target nor an independently optimized teacher.

The input tokens combine two patch embeddings,
\begin{equation}
 h_0=E_x(x_\sigma)+E_d(d).
\label{eq:embedding}
\end{equation}
$E_d$ is a separate convolution with the same patch geometry as $E_x$, initialized with zero weights and bias. This preserves the pretrained model's initial function while allowing the newly introduced pathway to use its own learning rate. The backbone, action embedder, and draft embedder are trainable; the autoencoder and text encoder remain frozen. Stopping gradients through the first pass does not freeze the shared backbone: it is updated through Eq.~\ref{eq:loss}. Feedback-enabled updates add one detached forward pass, giving $1+p=1.5$ expected network calls per update at $p=0.5$. Algorithm~\ref{alg:training} summarizes one training update.

\subsection{Refinement-Only Late Adapter}
\label{sec:adapter}
The draft and refinement passes share the entire transformer, but their inputs differ: one must propose a future without feedback, while the other can use a clean estimate. We provide the refinement role with a small residual transformation after block 30, immediately before the shared prediction head. For a token feature $h\in\mathbb{R}^{D}$,
\begin{align}
 q(h)&=\operatorname{GELU}(W_{\rm down}\operatorname{LN}(h)+b_{\rm down}),\notag\\
 A_\phi(h)&=W_{\rm up}q(h)+b_{\rm up}.
\label{eq:adapter}
\end{align}
\begin{equation}
 h' = h+rA_\phi(h).
\end{equation}
The LayerNorm is non-affine, $D=3072$, and the bottleneck width is 128. The up-projection and its bias start at zero, so attaching the adapter initially leaves the model output unchanged. The module contains 789,632 parameters.

The adapter is disabled for draft generation and for training examples without a draft. At inference it is enabled after the first clean estimate becomes available. It adds a tokenwise transformation within an existing network call. The model's output is a velocity prediction trained by Eq.~\ref{eq:loss}. The adapter gives draft-conditioned prediction a trainable correction at the final shared representation. With 789,632 parameters against a 5B backbone, its effect comes from where and when it acts rather than from its size: it is a role-specific residual path at the input of the shared head, applied to every token only in the refinement role. The role is also visible to the network through the draft input, since a zero draft marks the proposal role; the adapter makes this distinction explicit at the final representation, so the shared blocks serve both roles with one function and the correction specific to refinement is confined to a dedicated path. When the same adapter is active in both roles, one transformation has to serve proposal and refinement at once and also alters the draft that the refinement pass receives; restricting it to refinement leaves the draft pass identical to the shared prediction path and lets the transformation specialize to correcting a draft. Comparing against self-conditioning without the adapter measures the value of this role-specific path; an adapter active in both roles additionally controls for role assignment.

\subsection{Draft-Reuse Iterative Refinement}
\label{sec:inference}
This stage executes prediction feedback across sampling queries~\cite{analogbits}, using the unchanged UniPC solver~\cite{unipc}.

\paragraph{Initialization.} Initialize future latents from Gaussian noise, pin the boundary latent, and set $d_1=0$. The first query predicts without a draft and with the adapter disabled.

\paragraph{Recursive refinement.} Each query receives the updated noisy state, the actions, and the previous clean estimate. At query $k$, compute
\begin{align}
 v_k&=v_{\theta,\phi}(x_k,\tau_k,a,c,d_k;\mathbb 1[k>1]),\\
 \hat x_{0,k}&=x_k-\sigma_kv_k,
 \qquad \hat x_{0,k}^{(0)}\leftarrow b,\\
 x_{k+1}&=\operatorname{UniPCUpdate}(x_k,v_k,\text{solver history}),\\
 d_{k+1}&=\hat x_{0,k}.
\label{eq:inference}
\end{align}
The boundary is repinned after the solver update. Queries $2,\ldots,K$ use the incoming draft and enable the adapter, where $K$ is the NFE per chunk. After query $K$, the final solver state is decoded into the predicted video.

\paragraph{Fixed-call execution.} A single velocity prediction supplies both the solver update and the next clean draft. With action guidance scale 1, a $K$-query sample therefore uses $K$ denoiser calls and $K-1$ adapter activations. No additional draft forward pass is needed during sampling. The same model is evaluated at \nfe{4} and \nfe{6}, the two budgets of its training grid (other budgets in Sec.~\ref{sec:budget_time}). Training uses a same-state draft; inference uses the previous query's estimate.

\paragraph{Chunked rollout.} For a 33-frame rollout, generate two 17-frame chunks sharing one boundary. The decoded last frame of the first chunk becomes the observed frame for the second, whose action sequence contains the remaining 16 controls preceded by a zero initial action. The solver and draft state restart for the second chunk. The complete rollout requires $2K$ calls, with the second chunk conditioned on a generated observation; we report each chunk separately.

\input{tables/algorithm}

%% file: figures/method.tex
\begin{figure*}[t]
\centering
\includegraphics[width=\textwidth]{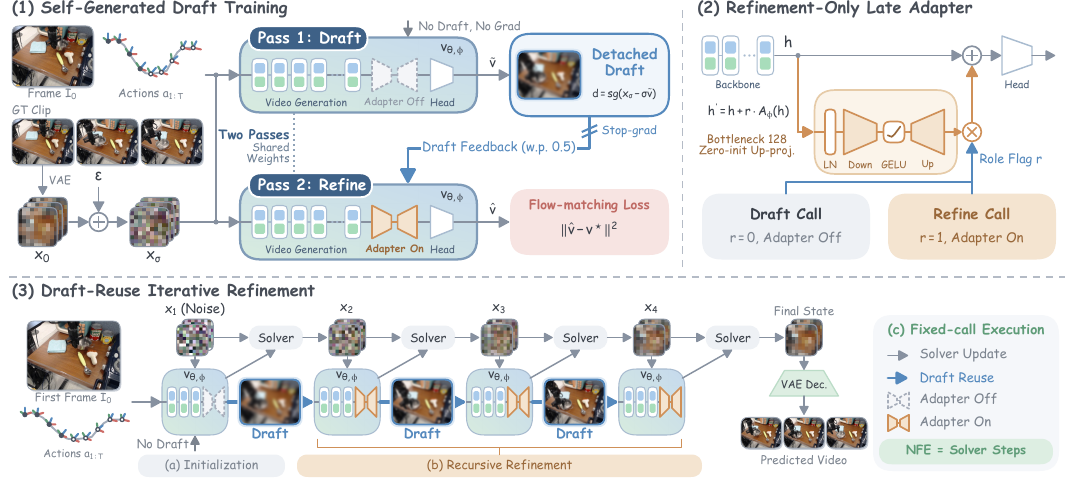}
\caption{\textbf{Overview of \method.} (1) \textbf{Self-Generated Draft Training:} on feedback-enabled updates (probability 0.5), two passes share weights and the noisy state. The first runs without gradients and produces a detached draft $d=\sg(x_\sigma-\sigma\tilde v)$ for the supervised second pass. (2) \textbf{Refinement-Only Late Adapter:} a bottleneck with a zero-initialized up-projection sits before the shared prediction head; the role flag $r$ disables it when no draft is available (draft call) and enables it for draft-conditioned (refine) calls. (3) \textbf{Draft-Reuse Iterative Refinement:} inference starts without a draft, then reuses each clean estimate while the solver advances the noisy state, so each solver step costs one network call. Draft training and draft reuse are stages of the same self-conditioning mechanism.}
\label{fig:method}
\end{figure*}

%% file: tables/algorithm.tex
\begin{algorithm}[t]
\caption{One \method\ training update}
\label{alg:training}
\begin{algorithmic}[1]
\Require Clean video latents $x_0$, boundary $b$, actions $a$, context $c$
\State Sample $(\sigma,\tau)\sim q_{\rm grid}$ (Sec.~\ref{sec:eval}); draw $\epsilon\sim\mathcal N(0,I)$
\State $x_\sigma\gets(1-\sigma)x_0+\sigma\epsilon$; $x_\sigma^{(0)}\gets b$
\State $d\gets0$; draw $z\sim\operatorname{Bernoulli}(0.5)$
\If{$z=1$}
\State Without gradients, $\tilde v\gets v_{\theta,\phi}(x_\sigma,\tau,a,c,0;0)$
\State $d\gets\sg(x_\sigma-\sigma\tilde v)$; $d^{(0)}\gets b$
\EndIf
\State $\hat v\gets v_{\theta,\phi}(x_\sigma,\tau,a,c,d;z)$
\State Update shared network and active adapter using Eq.~\ref{eq:loss}
\end{algorithmic}
\end{algorithm}

%% file: sections/4_experiments.tex
\section{Experiments}
\label{sec:experiments}
\method\ is self-conditioned training with the late adapter active only in the refinement role (Sec.~\ref{sec:adapter}). Its controls are Ordinary, the same backbone fine-tuned without draft pathway or adapter; SC, self-conditioned training without adapter; and SC+Adapter, self-conditioned training with the same adapter active in both roles, which controls for role assignment. All variants share the action representation, data, schedule and training-noise distribution, and every self-conditioned variant is sampled with draft reuse. The main configuration uses \nfe{4} per 17-frame chunk.

\subsection{Datasets and evaluation}
\label{sec:eval}
We evaluate on the official test splits: 2,946 windows from 1,725 Bridge episodes~\cite{bridge} and 5,243 windows from 4,529 Language Table episodes~\cite{language}. We report PSNR, SSIM~\cite{wang2004ssim}, LPIPS~\cite{lpips} and FVD~\cite{fvd}. Since each test window is driven by the recorded actions, agreement with the ground-truth future also reflects how well the prediction follows them. More experiments on motion regions and long-horizon prediction are in Appendix~\ref{sec:completed_motion}.

\paragraph{Training and conditioning.}
Models start from official Wan2.2-TI2V-5B weights and are trained for 300k iterations; training length is reported in iterations at an equivalent batch size of eight (samples seen divided by eight). Bridge training uses 32,768 windows from 25,113 episodes; Language Table uses one window from each of 32,768 episodes. Each \method\ or Wan control rollout receives one current RGB image and the future actions. Bridge uses task text; Language Table uses empty text. Clips have 17 frames at $256\times320$. Backbone/new-parameter learning rates are $10^{-6}/10^{-5}$; feedback probability is 0.5. Training uses eight H100 GPUs.

\input{tables/main_table1_external_native}
\input{tables/main_table2_equal_calls}
\paragraph{Baselines.}
IRASim~\cite{irasim} and ORV~\cite{yang2026orv4doccupancycentricrobot} retain their released weights and action-only interfaces. IRASim runs at its native resolution with 59 denoiser calls. The official 2B ORV model receives the complete current image and uses 50 calls; no Language Table model is released. RoboMaster~\cite{fu2026learningvideogenerationrobotic} receives additional inputs: object masks, robot--object image-plane trajectories and interaction-phase annotations, where \method\ receives only low-dimensional future robot actions. Its row is author-reported on a different test set (214 curated Bridge clips, 37 frames, $480\times640$) with 50 sampling steps. VDM~\cite{ho2022videodiffusionmodels} and LVDM~\cite{he2023latentvideodiffusionmodels} scores are those reported with IRASim; their times are measured here on the released models. Details are in Appendix~\ref{sec:appendix_baselines}.

\paragraph{Training-noise distribution.}
\label{sec:noise}
\method\ draws its training noise from the levels that the UniPC sampler visits at \nfe{4} or \nfe{6} (probabilities 0.4/0.6, then with equal probability within the schedule), under the standard flow objective; Table~\ref{tab:grid} lists them. We call this the query grid and train every model on it.

\subsection{Comparison with existing methods}
\label{sec:joint}
\input{figures/external_comparison}
At \nfe{4}, \method\ obtains PSNR 25.02\,dB, SSIM 0.853, LPIPS 0.062 and FVD 22.28 on Bridge (Table~\ref{tab:external_native}), improving all four displayed metrics over IRASim and over ORV. On Language Table, it improves three of the four metrics, SSIM (0.894 versus 0.879), LPIPS (0.059 versus 0.065) and FVD (18.94 versus 62.23), at 0.31\,s instead of 13.48\,s per chunk. The higher PSNR of IRASim (28.70) reflects per-pixel agreement, which favors smooth predictions, as the sampling-budget study in Sec.~\ref{sec:budget_time} also shows, whereas SSIM, LPIPS and FVD measure structure, perceptual similarity and temporal statistics, where \method\ leads by a wide margin. It also has higher SSIM and lower FVD than the reported LVDM results on both datasets, and is better than VDM on all three reported metrics. Figure~\ref{fig:external_candidate} shows selected examples: in \method's predictions the manipulated objects stay present with recognizable shape and color while they are moved, lifted, pushed and set upright, whereas each baseline misplaces, deforms or loses an object. Further qualitative examples are provided in Figures~\ref{fig:qualitative_supp_bridge} and~\ref{fig:qualitative_supp_lt}.

\paragraph{Equal call budgets.}
Table~\ref{tab:equal_calls} samples the released IRASim and ORV weights at \nfe{4} and \nfe{6}, using the samplers described in Appendix~\ref{sec:appendix_baselines}. At equal NFE, \method\ improves all four metrics over both baselines on Bridge, and SSIM, LPIPS and FVD over IRASim on Language Table. At \nfe{4} its Bridge FVD is 22.28, against 84.49 for IRASim and 75.22 for ORV, and across the table its chunk time is 2.2--4.1$\times$ shorter than the baseline at the same NFE. With fewer calls, IRASim keeps PSNR within 0.17\,dB of its 59-call result, while its FVD rises from 26.63 to 84.49 on Bridge and from 65.51 to 106.97 on the Language Table subset.

\paragraph{Object appearance consistency.}
\input{tables/main_table3_object_appearance}
Table~\ref{tab:object_appearance} measures how stable the appearance of individual objects is in the predictions of the main comparison. SAM~2~\cite{ravi2024sam2segmentimages} segmentation of the ground-truth first frame gives up to six objects per window, and the same prompts track these objects through each prediction. Appearance drift is the LPIPS between the crop of an object in a future frame and its crop in the first frame; flicker is the LPIPS between its crops in consecutive frames; both are averaged over frames, objects and windows, on 1,964 Bridge and 3,496 Language Table test windows at equal spacing. \method\ has the lowest drift and flicker on both datasets, with 4 calls against 59 for IRASim and 50 for ORV: the manipulated objects keep their appearance while they are moved.

\subsection{Ablation studies}
\label{sec:wan_ablation}

\input{tables/main_table4_matched_final}

Table~\ref{tab:matched_final} ablates the framework at the full 300k-iteration budget. Within each dataset, all variants share initialization, training data, seed, learning rates and training length. Relative to Ordinary, SC adds the draft pathway and self-conditioned training, SC+Adapter adds the late adapter in both roles, and \method\ restricts the adapter to the refinement role.

\paragraph{Component chain.}
Every step of the chain improves PSNR, SSIM, LPIPS and FVD on both datasets. Self-conditioning raises PSNR over Ordinary by 0.27\,dB on Bridge and 0.21\,dB on Language Table and lowers FVD by 0.59 and 1.10, and the adapter adds 0.16\,dB and 0.20\,dB. A feedback-path control in Table~\ref{tab:core} attributes the self-conditioning gain to the learned draft rather than to the added input pathway, and the chain repeated at shorter training budgets in Appendix~\ref{sec:module_training} gives the same ordering.

\paragraph{Role-specific adapter.}
Restricting the adapter to the refinement role adds a further 0.24\,dB on Bridge and 0.18\,dB on Language Table over the adapter active in both roles and lowers LPIPS and FVD. The adapter active in both roles also modifies the draft that the refinement pass receives and serves two tasks with one transformation, whereas the refinement-only adapter leaves the draft pass unchanged and specializes to correcting a draft (Sec.~\ref{sec:adapter}). In total, \method\ improves PSNR over Ordinary by 0.67\,dB and 0.59\,dB, raises SSIM by 0.041 and 0.012, lowers LPIPS by 0.007 and 0.006 and reduces FVD by 1.50 and 2.28.

\paragraph{Two-chunk rollouts.}
On 33-frame two-chunk rollouts (128 Bridge and 512 Language Table cases), \method\ improves second-chunk PSNR by 1.03\,dB on Bridge and 0.52\,dB on Language Table. Its error over the pixels that change between frames is also lower on both datasets, including the highest-motion quartile of test windows. More experiments on action accuracy are in Appendix~\ref{sec:action_consistency}.

\subsection{Sampling budget and inference time}
\label{sec:budget_time}
\input{tables/main_table5_sampling_budget_fulltest}
\input{tables/main_table6_latency}
Table~\ref{tab:sampling_budget_fulltest} evaluates \method\ from \nfe{2} to \nfe{8}: PSNR varies by 0.21\,dB on Bridge and 0.32\,dB on Language Table, while FVD falls from 24.88 to 19.91 and from 25.69 to 14.18. PSNR and SSIM are highest at \nfe{2}, LPIPS is nearly flat, and FVD improves steadily with more calls: additional calls add detail and temporal structure that the distribution-level metric rewards, while per-pixel agreement stays within 0.21\,dB and 0.32\,dB. \nfe{4} is the smallest budget at which LPIPS reaches its minimum on both datasets and FVD has gained most of its improvement. On the 500-window subset in Table~\ref{tab:k_sweep_subset}, \method\ has higher mean PSNR than its matched control at \nfe{2}, \nfe{4} and \nfe{6}, and the control shows the same trend across budgets. At \nfe{4} the ratio of predicted to ground-truth sharpness is 0.97 for both models on Bridge and 0.95 for both on Language Table. Decoding the estimate after each of the four calls, LPIPS falls from 0.069 to 0.062 and FVD from 31.81 to 22.28 on Bridge, and from 0.067 to 0.059 and from 46.70 to 20.42 on 2,622 Language Table windows, as shown in Table~\ref{tab:per_call_estimates}.

Table~\ref{tab:latency} reports generation time, with each method at its native resolution and frame count. At \nfe{4}, \method\ generates a chunk in 0.31\,s on both datasets, 25.6$\times$ faster than IRASim and 12.6$\times$ faster than ORV on Bridge and 43.5$\times$ faster than IRASim on Language Table. LVDM, the fastest baseline, is 7.0$\times$ and 10.5$\times$ slower.

\subsection{Training pass depth}
\label{sec:depth}
\input{tables/main_table7_npass_cost}
Training depth $N$ uses $N-1$ detached draft passes before one supervised pass; \method\ uses $N=2$. Table~\ref{tab:npass_cost} compares two, three and four passes at matched iterations and at matched training time, on 256 Bridge test cases at \nfe{10}. At matched iterations and relative to two passes, three passes change PSNR by $+0.03$\,dB for 10.3\% more training time, and four passes by $+0.01$\,dB for 21.8\% more. At matched training time, the three depths differ by at most 0.02\,dB, which supports $N=2$. Table~\ref{tab:npass_official} reports further training lengths.

%% file: tables/main_table1_external_native.tex
\begin{table}[t]
\centering\small
\setlength{\tabcolsep}{2.8pt}\renewcommand{\arraystretch}{1.15}
\caption{\textbf{Main comparison.} The RoboMaster row is reported in~\cite{fu2026learningvideogenerationrobotic} and the VDM and LVDM rows in~\cite{irasim}; all other rows are evaluated on our complete test sets, and bold marks the best displayed value among them in each block. Our method is named in bold. -- is not reported.}
\label{tab:external_native}
\begin{tabular}{@{}llcrrrr@{}}
\toprule
Dataset & Method & NFE & PSNR$\uparrow$ & SSIM$\uparrow$ & LPIPS$\downarrow$ & FVD$\downarrow$\\
\midrule
\multirow{6}{*}{Bridge} & VDM & 59 & 18.52 & 0.741 & -- & 127.25\\
 & LVDM & 59 & 23.55 & 0.810 & -- & 35.06\\
 & IRASim & 59 & 24.87 & 0.810 & 0.070 & 26.63\\
 & ORV & 50 & 23.57 & 0.832 & 0.117 & 41.73\\
 & RoboMaster & 50 & 21.55 & 0.803 & -- & 147.31\\
 & \textbf{TwoPass} & 4 & \textbf{25.02} & \textbf{0.853} & \textbf{0.062} & \textbf{22.28}\\
\midrule
\multirow{4}{*}{\shortstack[l]{Language\\Table}} & VDM & 59 & 23.07 & 0.857 & -- & 136.56\\
 & LVDM & 59 & 28.25 & 0.889 & -- & 24.34\\
 & IRASim & 59 & \textbf{28.70} & 0.879 & 0.065 & 62.23\\
 & \textbf{TwoPass} & 4 & 26.43 & \textbf{0.894} & \textbf{0.059} & \textbf{18.94}\\
\bottomrule
\end{tabular}
\end{table}

%% file: tables/main_table2_equal_calls.tex
\begin{table}[t]
\centering\small
\setlength{\tabcolsep}{1.6pt}\renewcommand{\arraystretch}{1.15}
\caption{\textbf{Baselines at equal call budgets.} Released IRASim and ORV weights sampled with fewer denoiser calls. Bridge uses the complete test set; on Language Table, the IRASim rows use 2,000 equally spaced test windows and the TwoPass rows the complete test set, as in Table~\ref{tab:external_native}. Time is the median generation time per chunk in seconds; the 59- and 50-call baseline rows and the TwoPass rows use the timings of Table~\ref{tab:latency}.}
\label{tab:equal_calls}
\begin{tabular}{@{}llcrrrrr@{}}
\toprule
Dataset & Method & NFE & PSNR$\uparrow$ & SSIM$\uparrow$ & LPIPS$\downarrow$ & FVD$\downarrow$ & Time$\downarrow$\\
\midrule
\multirow{8}{*}{Bridge} & IRASim & 59 & 24.87 & 0.810 & 0.070 & 26.63 & 7.94\\
 & IRASim & 6 & 24.71 & 0.799 & 0.085 & 65.53 & 0.95\\
 & IRASim & 4 & 24.70 & 0.792 & 0.096 & 84.49 & 0.69\\
 & ORV & 50 & 23.57 & 0.832 & 0.117 & 41.73 & 3.90\\
 & ORV & 6 & 23.13 & 0.818 & 0.122 & 50.13 & 1.09\\
 & ORV & 4 & 24.47 & 0.848 & 0.118 & 75.22 & 0.95\\
 & \textbf{TwoPass} & 6 & 24.95 & 0.852 & \textbf{0.062} & \textbf{21.14} & 0.39\\
 & \textbf{TwoPass} & 4 & \textbf{25.02} & \textbf{0.853} & \textbf{0.062} & 22.28 & \textbf{0.31}\\
\midrule
\multirow{5}{*}{\shortstack[l]{Language\\Table}} & IRASim & 59 & 28.74 & 0.880 & 0.065 & 65.51 & 13.48\\
 & IRASim & 6 & \textbf{28.76} & 0.869 & 0.096 & 87.43 & 1.60\\
 & IRASim & 4 & 28.65 & 0.862 & 0.109 & 106.97 & 1.16\\
 & \textbf{TwoPass} & 6 & 26.34 & \textbf{0.894} & \textbf{0.059} & \textbf{17.85} & 0.39\\
 & \textbf{TwoPass} & 4 & 26.43 & \textbf{0.894} & \textbf{0.059} & 18.94 & \textbf{0.31}\\
\bottomrule
\end{tabular}
\end{table}

%% file: figures/external_comparison.tex
\begin{figure*}[t]
\centering
\includegraphics[width=\textwidth]{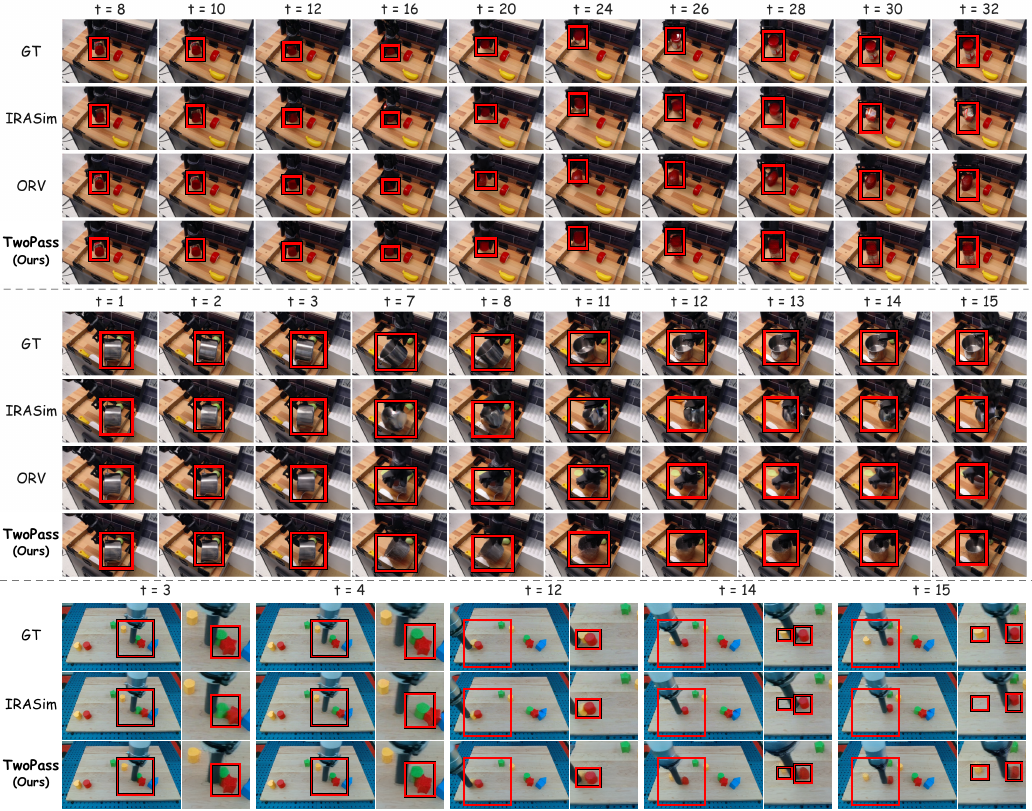}
\caption{\textbf{Selected predictions on Bridge and Language Table.} Top: placing a basil bottle upright, frames 8--32, where each method continues from its own predictions beyond the first chunk. Middle: moving a metal pot and setting it upright, frames 1--3, 7, 8 and 11--15. Bridge rows show GT, IRASim, ORV and TwoPass (\nfe{4}) on the common ORV crop. Bottom: Language Table, with GT, IRASim and TwoPass (\nfe{4}) at frames 3, 4, 12, 14 and 15, showing successive pushes of several blocks; each frame pairs the full view with a zoom whose region follows the interaction. Red boxes use coordinates derived from GT and shared across methods; the outline in the full view marks the zoom region. Methods share the initial observation and actions.}
\label{fig:external_candidate}
\end{figure*}

%% file: tables/main_table3_object_appearance.tex
\begin{table}[t]
\centering\small
\setlength{\tabcolsep}{7pt}\renewcommand{\arraystretch}{1.15}
\caption{\textbf{Object appearance consistency.} LPIPS between crops of tracked objects, on 1,964 Bridge and 3,496 Language Table test windows. TwoPass uses \nfe{4}.}
\label{tab:object_appearance}
\begin{tabular}{@{}llrr@{}}
\toprule
Dataset & Method & Appearance drift$\downarrow$ & Flicker$\downarrow$\\
\midrule
\multirow{3}{*}{Bridge} & IRASim & 0.173 & 0.071\\
 & ORV & 0.181 & 0.060\\
 & \textbf{TwoPass} & \textbf{0.160} & \textbf{0.056}\\
\midrule
\multirow{2}{*}{\shortstack[l]{Language\\Table}} & IRASim & 0.186 & 0.093\\
 & \textbf{TwoPass} & \textbf{0.165} & \textbf{0.071}\\
\bottomrule
\end{tabular}
\end{table}

%% file: tables/main_table4_matched_final.tex
\begin{table}[t]
\centering\small
\setlength{\tabcolsep}{5pt}
\renewcommand{\arraystretch}{1.15}
\caption{\textbf{Matched training controls, \nfe{4}.} 300k iterations, complete test sets. Ordinary: no draft pathway, no adapter; SC: self-conditioned training; SC+Adapter: adapter active in both roles; TwoPass: adapter active in the refinement role only.}
\label{tab:matched_final}
\begin{tabular}{@{}llrrrr@{}}
\toprule
Dataset & Variant & PSNR$\uparrow$ & SSIM$\uparrow$ & LPIPS$\downarrow$ & FVD$\downarrow$\\
\midrule
\multirow{4}{*}{Bridge}
 & Ordinary & 24.35 & 0.812 & 0.069 & 23.78\\
 & SC & 24.62 & 0.829 & 0.065 & 23.19\\
 & SC+Adapter & 24.78 & 0.844 & 0.064 & 22.91\\
 & \textbf{TwoPass} & \textbf{25.02} & \textbf{0.853} & \textbf{0.062} & \textbf{22.28}\\
\midrule
\multirow{4}{*}{\shortstack[l]{Language\\Table}}
 & Ordinary & 25.84 & 0.882 & 0.065 & 21.22\\
 & SC & 26.05 & 0.887 & 0.062 & 20.12\\
 & SC+Adapter & 26.25 & 0.891 & 0.061 & 19.57\\
 & \textbf{TwoPass} & \textbf{26.43} & \textbf{0.894} & \textbf{0.059} & \textbf{18.94}\\
\bottomrule
\end{tabular}
\end{table}

%% file: tables/main_table5_sampling_budget_fulltest.tex
\begin{table}[t]
\centering\small
\setlength{\tabcolsep}{4pt}\renewcommand{\arraystretch}{1.0}
\caption{\textbf{Sampling budget.} TwoPass on the complete test sets. Best in bold, second best underlined.}
\label{tab:sampling_budget_fulltest}
\begin{tabular}{@{}llrrrr@{}}
\toprule
Dataset & NFE & PSNR$\uparrow$ & SSIM$\uparrow$ & LPIPS$\downarrow$ & FVD$\downarrow$\\
\midrule
\multirow{5}{*}{Bridge} & 2 & \textbf{25.14} & \textbf{0.854} & 0.064 & 24.88\\
 & 3 & \underline{25.08} & \underline{0.853} & \underline{0.063} & 23.09\\
 & 4 & 25.02 & \underline{0.853} & \textbf{0.062} & 22.28\\
 & 6 & 24.95 & 0.852 & \textbf{0.062} & \underline{21.14}\\
 & 8 & 24.93 & 0.851 & \textbf{0.062} & \textbf{19.91}\\
\midrule
\multirow{5}{*}{\shortstack[l]{Language\\Table}} & 2 & \textbf{26.63} & \textbf{0.894} & \underline{0.060} & 25.69\\
 & 3 & \underline{26.51} & \textbf{0.894} & \textbf{0.059} & 20.57\\
 & 4 & 26.43 & \textbf{0.894} & \textbf{0.059} & 18.94\\
 & 6 & 26.34 & \textbf{0.894} & \textbf{0.059} & \underline{17.85}\\
 & 8 & 26.31 & \underline{0.893} & \underline{0.060} & \textbf{14.18}\\
\bottomrule
\end{tabular}
\end{table}

%% file: tables/main_table6_latency.tex
\begin{table}[t]
\centering\small
\setlength{\tabcolsep}{2.5pt}\renewcommand{\arraystretch}{1.0}
\caption{\textbf{Median generation time per chunk} in seconds on a single H100 GPU at batch one, each method at its native resolution and frame count. Denoiser-only times are given in Table~\ref{tab:latency_full} and timing scopes in Appendix~\ref{sec:appendix_baselines}.}
\label{tab:latency}
\begin{tabular}{@{}lcr@{\hspace{8pt}}lcr@{}}
\toprule
\multicolumn{3}{c}{Bridge} & \multicolumn{3}{c}{Language Table}\\
\cmidrule(r){1-3}\cmidrule(l){4-6}
Method & NFE & Time$\downarrow$ & Method & NFE & Time$\downarrow$\\
\midrule
VDM & 59 & 22.41 & VDM & 59 & 10.12\\
LVDM & 59 & 2.17 & LVDM & 59 & 3.27\\
IRASim & 59 & 7.94 & IRASim & 59 & 13.48\\
ORV & 50 & 3.90 & -- & -- & --\\
\textbf{TwoPass} & 6 & 0.39 & \textbf{TwoPass} & 6 & 0.39\\
\textbf{TwoPass} & 4 & \textbf{0.31} & \textbf{TwoPass} & 4 & \textbf{0.31}\\
\bottomrule
\end{tabular}
\vspace{-8pt}
\end{table}

%% file: tables/main_table7_npass_cost.tex
\begin{table}[t]
\centering\small
\setlength{\tabcolsep}{4pt}\renewcommand{\arraystretch}{1.0}
\caption{\textbf{Training pass depth $N$.} Bridge, 256 test cases, \nfe{10}, 16-frame training clips. Time (h) is H100 GPU-hours divided by eight.}
\label{tab:npass_cost}
\begin{tabular}{@{}llrrrr@{}}
\toprule
Matched & $N$ & PSNR$\uparrow$ & SSIM$\uparrow$ & LPIPS$\downarrow$ & Time (h)$\downarrow$\\
\midrule
\multirow{3}{*}{Iterations} & 2 & 25.92 & \textbf{0.883} & \textbf{0.054} & \textbf{2.61}\\
 & 3 & \textbf{25.95} & \textbf{0.883} & \textbf{0.054} & 2.88\\
 & 4 & 25.93 & \textbf{0.883} & \textbf{0.054} & 3.18\\
\midrule
\multirow{3}{*}{Time} & 2 & 25.92 & \textbf{0.883} & \textbf{0.054} & \textbf{2.61}\\
 & 3 & \textbf{25.93} & 0.882 & \textbf{0.054} & \textbf{2.61}\\
 & 4 & 25.91 & \textbf{0.883} & \textbf{0.054} & \textbf{2.61}\\
\bottomrule
\end{tabular}
\vspace{-8pt}
\end{table}

%% file: sections/5_conclusion.tex
\section{Conclusion}
We presented \method, a self-conditioned training-and-inference framework for action-conditioned video prediction. Self-Generated Draft Training and a Refinement-Only Late Adapter teach one predictor to propose a future and to refine its own draft under the standard flow-matching objective, and Draft-Reuse Iterative Refinement runs it with as few as 4 NFE per video chunk. On Bridge and Language Table, \method\ achieves state-of-the-art results with 4 NFE, far fewer than the 50--59 calls of existing predictors.

%% file: sections/6_appendix.tex
\section{Implementation Details}
\label{sec:appendix_impl}
\paragraph{Parameter and gradient paths.}
The pretrained backbone is Wan2.2-TI2V-5B~\cite{wan}. Its 48-channel video latents use a temporal stride of four with a separately encoded first frame. We add an action embedder and, for feedback variants, a separate convolutional draft patch embedding. The action MLP has hidden widths 1024 and 3072 before a projection to six 3072-dimensional modulation vectors. The final action projection and the entire draft embedding are initialized to zero. The bottleneck-128 adapter uses a non-affine LayerNorm and two biased linear layers, giving $3072\cdot128+128+128\cdot3072+3072=789{,}632$ parameters.

All backbone parameters are updated through the supervised pass. During draft construction, autograd is disabled; the clean estimate is detached before re-entering the model. Adapter parameters receive supervised gradients only on active-feedback updates. This is role-specific computation, rather than a frozen draft model or a second trainable backbone. Training uses AdamW with $\beta=(0.9,0.95)$ and no weight decay, gradient-norm clipping at 1.0 and bfloat16 mixed precision, without EMA; the action embedder, draft embedding and adapter use the new-parameter learning rate.

\paragraph{Action alignment.}
For a 17-frame window, actions comprise an initial zero followed by 16 transition controls. The four future latent groups average controls 1--4, 5--8, 9--12, and 13--16. Bridge~\cite{bridge} scales its first six action coordinates by 20 and leaves the seventh coordinate unchanged. Language Table~\cite{language} supplies two planar displacement coordinates, scales each by 20, and sets the other five coordinates to zero. The action representation is shared by all ablation variants within a dataset.

\paragraph{Chunk boundary.}
The first chunk's decoded terminal image is clipped, quantized to 8-bit RGB, and re-encoded as the second chunk's boundary condition. The first output frame is an autoencoder reconstruction. The second chunk uses a fresh solver and feedback state and the remaining controls. Metrics exclude the initial observation.

\section{Baseline and Timing Details}
\label{sec:appendix_baselines}

\paragraph{External baselines.}
The 59 denoiser calls of IRASim~\cite{irasim} correspond to 50 nominal PNDM~\cite{liu2022pseudonumericalmethodsdiffusion} steps. The official 2B ORV model receives the complete current image resized to $384\times480$ (height $\times$ width), without cropping, generates 17 frames, and uses guidance scale one and empty text. The RoboMaster~\cite{fu2026learningvideogenerationrobotic} sampler uses 50 DDIM~\cite{song2022denoisingdiffusionimplicitmodels} steps with guidance scale 6. NFE in all tables counts denoiser forward passes per chunk, and a forward pass that evaluates the conditional and unconditional branches as one batch counts once. Under this convention RoboMaster has NFE 50, corresponding to 100 single-branch evaluations, 50 conditional and 50 unconditional. The VDM~\cite{ho2022videodiffusionmodels} and LVDM~\cite{he2023latentvideodiffusionmodels} rows are the short-trajectory results reported with IRASim, whose authors trained and evaluated both baselines on the same test splits without reporting LPIPS; their NFE is the call count of the released evaluation pipeline. For the equal-call comparison, ORV keeps its scheduler with fewer steps, and IRASim uses the best of DDIM, DPM-Solver++~\cite{lu2025dpmsolverfastsolverguided} and UniPC~\cite{unipc}: UniPC on Bridge and DDIM on Language Table.

\paragraph{Inference-time measurement.}
The timing table in the main paper and Table~\ref{tab:latency_full} report the median over repeated runs on a single H100 GPU at batch one, after one warm-up. Total time includes text encoding and VAE encoding and decoding for \method\ and ORV; the IRASim timer covers denoising and VAE decoding but not its first-frame encoding. Model loading, scoring and output saving are excluded. IRASim generates 16 frames at $256\times320$ on Bridge and $288\times512$ on Language Table, ORV 17 frames at $384\times480$, and \method\ 17 frames at $256\times320$. VDM and LVDM generate 16 frames at the IRASim resolutions, except VDM at $144\times256$ on Language Table.
\input{tables/supp_table01_latency_full}

\section{Training-Noise Distribution}
\label{sec:appendix_grid}
Table~\ref{tab:grid} lists the noise levels and timesteps of the query grid. Each listed entry has equal probability, so the two levels that occur in both schedules (the highest and approximately 0.8333) are drawn twice as often. Each training example draws its level independently; the feedback draw is made once per update and is independent of the noise level. The terminal zero-noise value is excluded because it is not a denoiser query. No importance weighting is applied.

\input{tables/supp_table02_grid}

\section{Per-Call Estimates}
Table~\ref{tab:per_call_estimates} decodes the clean estimate after each call at \nfe{4}; the Language Table rows use every second test window. Later calls lower LPIPS and FVD, the latter from 31.81 to 22.28 on Bridge and from 46.70 to 20.42 on Language Table.
\input{tables/supp_table03_per_call_estimates}

\section{Sampling Budget and Matched Controls}
\label{sec:appendix_sampling}
Table~\ref{tab:k_sweep_subset} compares \method\ with its matched training controls at \nfe{2}, \nfe{4} and \nfe{6}, on 500 equally spaced test windows per dataset. \method\ has higher PSNR and SSIM and lower LPIPS than its control at every budget on both datasets. The PSNR gain grows with the budget, from 0.27 to 0.45\,dB on Bridge and from 0.35 to 0.63\,dB on Language Table, and SSIM is higher by 0.022--0.024 on Bridge and 0.011--0.012 on Language Table. LPIPS is lower by 0.0029--0.0048 on Bridge and 0.0058--0.0078 on Language Table, with the largest reduction at \nfe{2} on both datasets.
\input{tables/supp_table04_k_sweep_subset}

\section{Motion Regions and Long-Horizon Evaluation}
\label{sec:completed_motion}
The motion analysis uses the \nfe{4} predictions on the complete test sets. A pixel is marked as changing when the channel-mean absolute difference between adjacent ground-truth frames exceeds $8/255$. Table~\ref{tab:motion_regions} reports all test windows and the highest-motion quartiles: 736 Bridge windows from 612 episodes and 1,310 Language Table windows from 1,246 episodes. Copying the observed frame gives motion MSE 0.07449/0.05921 on Bridge/Language Table, versus \method's 0.02791/0.03580.
\input{tables/supp_table05_motion_regions}

The 33-frame study uses the same models at \nfe{4} with two autoregressive chunks, on 128 Bridge and 512 Language Table cases with sufficient episode length. Each method conditions its second chunk on its own generated boundary. Table~\ref{tab:long_horizon} reports \method\ minus SC on Bridge and minus ordinary Wan on Language Table; SSIM and LPIPS improve on both datasets for the second chunk and for the full future.

\input{tables/supp_table06_long_horizon}

\section{Action Consistency}
\label{sec:action_consistency}
This section measures how closely the predicted motion follows the actions: the positions of tracked objects against the ground truth, and whether the prediction made with the recorded actions is closer to the ground-truth future than predictions made with other actions. Controls are the matched models of the main paper, SC on Bridge and ordinary Wan on Language Table; all models use \nfe{4}. Significance refers to paired 95\% intervals for \method\ minus the other method from 2,000 episode-level bootstrap resamples, and subsets are taken at equal spacing from the test lists.

\paragraph{Object motion accuracy.}
SAM~2~\cite{ravi2024sam2segmentimages} segmentation of the ground-truth first frame gives up to six objects per window, and the same prompts track each object through the ground-truth video and through each prediction; the mask centroid is the object position. Moving objects are those whose ground-truth centroid moves by at least 10 pixels from the first frame. On Bridge, the moving object with a first-frame area of at least 1,500 pixels that travels farthest is taken as the arm and the remaining moving objects as manipulated objects; on Language Table all moving objects are reported together. Center error is the mean centroid distance to the ground truth over frames 1--15, displacement error the absolute difference between predicted and ground-truth travel distance, endpoint error the centroid distance in the last frame, and placement accuracy the fraction of objects whose last-frame centroid lies within 10 pixels of the ground truth; an object that vanishes from a prediction is scored at its first-frame position. Values are averaged over the objects of a window and then over windows, on the 958 of 1,000 equally spaced Bridge windows and the 1,432 of 1,500 Language Table windows that contain a moving object; the predictions are those of the main comparison. Table~\ref{tab:object_motion} reports the values and Table~\ref{tab:object_motion_diff} the paired differences. \method\ lowers all three errors over its matched control on both datasets, and raises placement accuracy by 7.3 points on Bridge and 6.3 points on Language Table; every one of these differences is significant. On Bridge it also has lower center, displacement and endpoint errors than ORV, over all moving objects, over the manipulated objects and over the arm. The static reference, which keeps every object at its first-frame position, is given for scale.

\paragraph{Action discrimination.}
For every window of the complete test sets, \method\ predicts the future twice from the same first frame, text and sampling noise: once with the recorded actions and once with alternative actions, either those of the next test window from a different episode (swapped) or all zeros (zeroed). A window counts as correct when the prediction made with the recorded actions is closer to the ground-truth future. Table~\ref{tab:action_discrimination} gives the percentages: the recorded actions are identified in 99.6\% of Bridge windows and 97.6\% of Language Table windows by LPIPS against swapped actions, and in at least 99.3\% and 96.4\% under both distances and both alternatives.

\paragraph{Candidate action ranking.}
On 300 equally spaced windows per dataset, each window receives eight candidate action sequences: the recorded one and seven drawn with a fixed seed from test windows of other episodes. The model generates one future per candidate, and the candidates are ranked by the distance of their prediction to the ground-truth future, under LPIPS, PSNR and the MSE over changed pixels, that is, ground-truth pixels that differ from the first frame by more than 0.1. Table~\ref{tab:action_ranking} reports the percentage of windows in which the recorded actions rank first, the mean reciprocal rank and the mean rank; chance level is 12.5\%, 0.34 and 4.5. The recorded actions rank first in 98.0--98.3\% of Bridge windows and 93.3--94.7\% of Language Table windows. Table~\ref{tab:ranking_margin} reports the margin by which the recorded actions beat the closest wrong candidate: \method\ separates the recorded actions from the best alternative by a wider margin than its matched control on both datasets and under all three distances, and every gain is significant. On subsets of 102 Bridge and 50 Language Table windows with the same candidates, IRASim with the full sampler of the main comparison ranks the recorded actions first in 98.0\% and 98.0\% of windows by PSNR, against 99.0\% and 98.0\% for \method; ranking the eight candidates of one window takes IRASim 63.5\,s on Bridge and 107.8\,s on Language Table, and \method\ 2.5\,s, eight generations at its measured 0.31\,s per chunk.

\paragraph{Fine-grained action ranking.}
On the same 300 windows, the eight candidates mix the recorded actions with the actions of another window in proportions of 0, 10, 20, 30, 50, 70, 85 and 100\%, so that the candidates deviate from the recorded actions by graded amounts. Table~\ref{tab:fine_ranking} reports the Spearman correlation between mixing proportion and prediction distance and the percentage of candidate pairs ordered correctly. Under every distance and on both datasets, \method\ orders the candidates at least as well as its matched control; on Language Table the gains in Spearman correlation under LPIPS and changed-pixel MSE and in pairwise accuracy under all three distances are significant, as is the pairwise gain under PSNR on Bridge. Table~\ref{tab:mix_curve} shows that the prediction error of both models rises monotonically with the deviation of the actions and that \method\ is more accurate for correct and nearly correct actions: at proportions up to 30\%, its PSNR is higher by 0.2--0.7\,dB on Bridge and 0.1--0.6\,dB on Language Table.
\input{tables/supp_table07_object_motion}
\input{tables/supp_table08_object_motion_diff}
\input{tables/supp_table09_action_discrimination}
\input{tables/supp_table10_action_ranking}
\input{tables/supp_table11_ranking_margin}
\input{tables/supp_table12_fine_ranking}
\input{tables/supp_table13_mix_curve}

\section{Additional Qualitative Results}
\label{sec:appendix_qualitative}
\begin{figure*}[t]
\centering
\includegraphics[width=\textwidth]{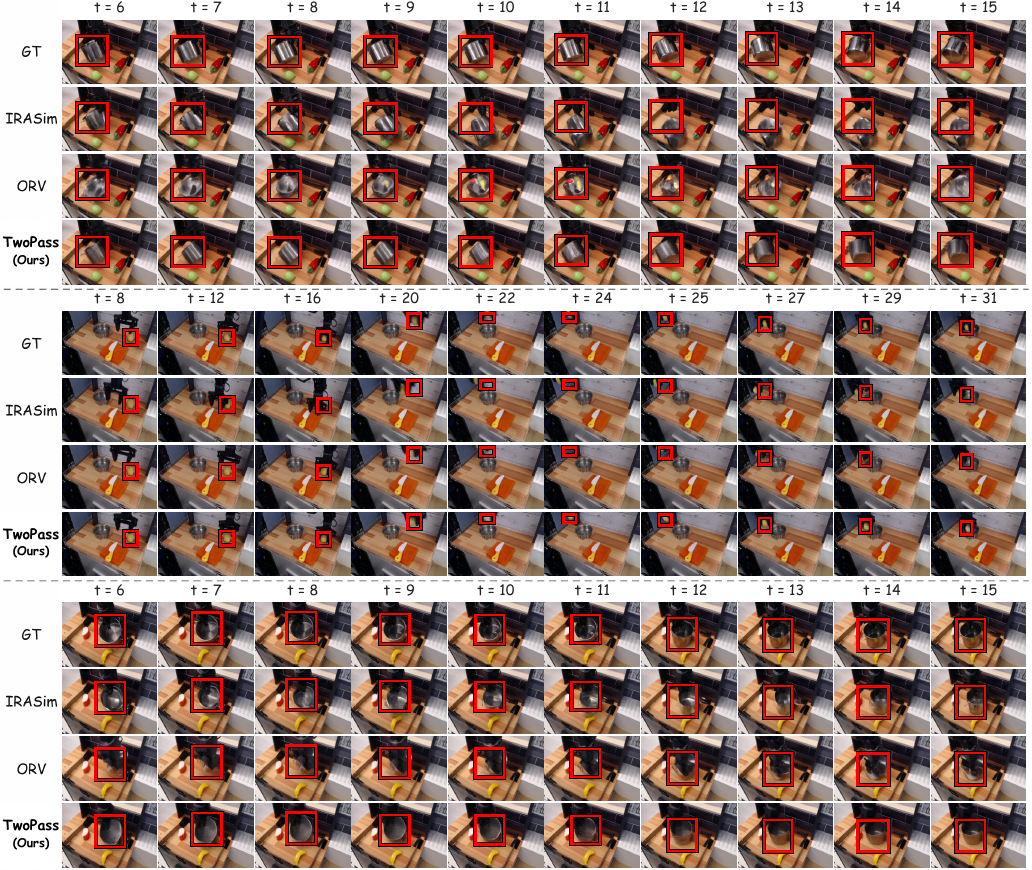}
\caption{\textbf{Additional predictions on Bridge.} Rows show GT, IRASim, ORV and TwoPass (\nfe{4}), at frames 6--15 in the top and bottom examples and at frames 8, 12, 16, 20, 22, 24, 25, 27, 29 and 31 in the middle example. Top: moving a blue cloth; TwoPass places the cloth closest to the ground truth, and both baselines place it with an offset. Middle: closing the flap of a box; in the last three frames TwoPass follows the closing flap, and the flap of both baselines remains partly open. Bottom: moving a metal pot and setting it upright; TwoPass stays close to the ground-truth position and size of the pot, IRASim's pot lags behind, and ORV's pot shrinks in the middle frames. Red boxes use coordinates derived from GT and shared across methods. Methods share the initial observation and actions.}
\label{fig:qualitative_supp_bridge}
\end{figure*}

\begin{figure*}[t]
\centering
\includegraphics[width=\textwidth]{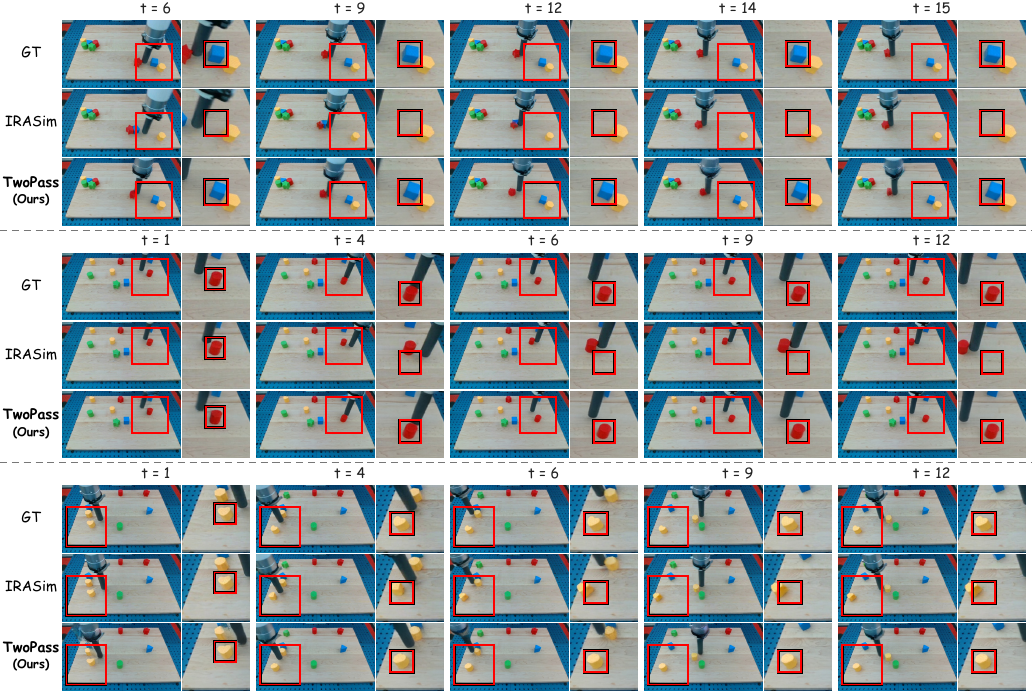}
\caption{\textbf{Additional predictions on Language Table.} Rows show GT, IRASim and TwoPass (\nfe{4}); each frame pairs the full view with a zoom on a region that is the same for all rows and frames of an example. Top (frames 6, 9, 12, 14 and 15): a blue block that stays in place; TwoPass keeps it there, and IRASim loses it. Middle (frames 1, 4, 6, 9 and 12): a red block pushed by the arm; TwoPass keeps the block at the ground-truth position throughout, and IRASim's block leaves the region after frame 4. Bottom (frames 1, 4, 6, 9 and 12): a yellow block pushed by the arm; TwoPass matches the ground-truth position and size of the block, and IRASim's block is displaced from frame 6 on. Red boxes use coordinates derived from GT and shared across methods; the outline in the full view marks the zoom region. Methods share the initial observation and actions.}
\label{fig:qualitative_supp_lt}
\end{figure*}

Figures~\ref{fig:qualitative_supp_bridge} and~\ref{fig:qualitative_supp_lt} show further examples on Bridge and Language Table, drawn in the same way as the comparison figure of the main paper. They cover translation, an articulated flap, reorientation and pushing.

\input{sections/module_studies}

%% file: tables/supp_table01_latency_full.tex
\begin{table}[t]
\centering\footnotesize
\setlength{\tabcolsep}{6pt}\renewcommand{\arraystretch}{1.1}
\caption{\textbf{Median generation time per chunk, total and denoiser-only,} on a single H100 GPU at batch one. Total includes text encoding and VAE encoding and decoding for TwoPass and ORV; the timing scope of each baseline is described above.}
\label{tab:latency_full}
\begin{tabular}{@{}llcrr@{}}
\toprule
Dataset & Method & NFE & Total (s)$\downarrow$ & Denoiser (s)$\downarrow$\\
\midrule
\multirow{6}{*}{Bridge} & VDM & 59 & 22.41 & 22.36\\
 & LVDM & 59 & 2.17 & 1.99\\
 & IRASim & 59 & 7.94 & 7.76\\
 & ORV & 50 & 3.90 & 2.43\\
 & \textbf{TwoPass} & 6 & 0.39 & 0.20\\
 & \textbf{TwoPass} & 4 & \textbf{0.31} & \textbf{0.13}\\
\midrule
\multirow{5}{*}{\shortstack[l]{Language\\Table}} & VDM & 59 & 10.12 & 10.08\\
 & LVDM & 59 & 3.27 & 2.98\\
 & IRASim & 59 & 13.48 & 13.19\\
 & \textbf{TwoPass} & 6 & 0.39 & 0.21\\
 & \textbf{TwoPass} & 4 & \textbf{0.31} & \textbf{0.14}\\
\bottomrule
\end{tabular}
\end{table}

%% file: tables/supp_table02_grid.tex
\begin{table}[t]
\centering\footnotesize
\setlength{\tabcolsep}{4pt}
\caption{\textbf{Query grid.} Noise levels $\sigma$ and timesteps $\tau$ that the UniPC~\cite{unipc} sampler (shift 5) visits at \nfe{4} and \nfe{6}, where NFE is the number of function evaluations per chunk. Each entry has probability 0.1 in training. The zero-noise terminal value is not sampled.}
\label{tab:grid}
\begin{tabular}{@{}crrrrrr@{}}
\toprule
 & \multicolumn{6}{c}{Query order}\\
NFE & 1 & 2 & 3 & 4 & 5 & 6\\
\midrule
4, $\sigma$ & .999960 & .937453 & .833278 & .624937 & -- & --\\
4, $\tau$ & 999 & 937 & 833 & 624 & -- & --\\
6, $\sigma$ & .999960 & .961494 & .909041 & .833278 & .714224 & .499940\\
6, $\tau$ & 999 & 961 & 909 & 833 & 714 & 499\\
\bottomrule
\end{tabular}
\end{table}

%% file: tables/supp_table03_per_call_estimates.tex
\begin{table}[t]
\centering\footnotesize
\setlength{\tabcolsep}{8pt}
\caption{\textbf{Estimates after each call.} TwoPass at \nfe{4}. Call $k$ scores the decoded clean estimate after the $k$-th denoiser call; call 4 is the model output. Language Table rows use 2,622 test windows.}
\label{tab:per_call_estimates}
\begin{tabular}{@{}llrr@{}}
\toprule
Dataset & Estimate & LPIPS$\downarrow$ & FVD$\downarrow$\\
\midrule
\multirow{4}{*}{Bridge} & Call 1 & 0.06940 & 31.81\\
 & Call 2 & 0.06474 & 26.06\\
 & Call 3 & 0.06344 & 24.07\\
 & \textbf{Call 4 (TwoPass)} & \textbf{0.06218} & \textbf{22.28}\\
\midrule
\multirow{4}{*}{\shortstack[l]{Language\\Table}} & Call 1 & 0.06671 & 46.70\\
 & Call 2 & 0.06127 & 29.57\\
 & Call 3 & 0.05981 & 24.40\\
 & \textbf{Call 4 (TwoPass)} & \textbf{0.05908} & \textbf{20.42}\\
\bottomrule
\end{tabular}
\end{table}

%% file: tables/supp_table04_k_sweep_subset.tex
\begin{table}[t]
\centering\footnotesize
\setlength{\tabcolsep}{4pt}
\caption{\textbf{Comparison with the matched training controls at equal NFE.}
TwoPass and its matched training control (300k iterations)
on 500 equally spaced test windows per dataset.
Sharp.\ is the mean gradient magnitude of the predicted
future frames divided by that of the ground truth.}
\label{tab:k_sweep_subset}
\begin{tabular}{@{}lclrrrr@{}}
\toprule
Dataset & NFE & Model & PSNR$\uparrow$ & SSIM$\uparrow$ & LPIPS$\downarrow$ & Sharp.\\
\midrule
\multirow{6}{*}{Bridge}
 & \multirow{2}{*}{2} & SC
 & 24.755 & 0.8285 & 0.06937 & 0.9558\\
 & & \textbf{TwoPass}
 & \textbf{25.023} & \textbf{0.8506}
 & \textbf{0.06462} & 0.9629\\
\addlinespace[2pt]
 & \multirow{2}{*}{4} & SC
 & 24.497 & 0.8271 & 0.06591 & 0.9652\\
 & & \textbf{TwoPass}
 & \textbf{24.912} & \textbf{0.8500}
 & \textbf{0.06304} & 0.9662\\
\addlinespace[2pt]
 & \multirow{2}{*}{6} & SC
 & 24.398 & 0.8249 & 0.06617 & 0.9669\\
 & & \textbf{TwoPass}
 & \textbf{24.844} & \textbf{0.8492}
 & \textbf{0.06277} & 0.9681\\
\midrule
\multirow{6}{*}{\shortstack[l]{Language\\Table}}
 & \multirow{2}{*}{2} & Ordinary
 & 26.469 & 0.8854 & 0.06496 & 0.9440\\
 & & \textbf{TwoPass}
 & \textbf{26.821} & \textbf{0.8966}
 & \textbf{0.05721} & 0.9481\\
\addlinespace[2pt]
 & \multirow{2}{*}{4} & Ordinary
 & 26.057 & 0.8839 & 0.06172 & 0.9533\\
 & & \textbf{TwoPass}
 & \textbf{26.626} & \textbf{0.8963}
 & \textbf{0.05588} & 0.9524\\
\addlinespace[2pt]
 & \multirow{2}{*}{6} & Ordinary
 & 25.917 & 0.8841 & 0.06241 & 0.9554\\
 & & \textbf{TwoPass}
 & \textbf{26.551} & \textbf{0.8959}
 & \textbf{0.05604} & 0.9536\\
\bottomrule
\end{tabular}
\end{table}

%% file: tables/supp_table05_motion_regions.tex
\begin{table}[t]
\centering\footnotesize
\setlength{\tabcolsep}{3pt}
\caption{\textbf{Motion-region MSE at 300k iterations, \nfe{4}.} Bridge control: SC; Language Table control: ordinary Wan. High motion is the top quartile by ground-truth visual change. Intervals are paired 95\% intervals for TwoPass minus control.}
\label{tab:motion_regions}
\begin{tabular}{@{}llrrr@{}}
\toprule
Dataset & Subset & Control & \textbf{TwoPass} & Difference interval\\
\midrule
\multirow{2}{*}{Bridge} & All & 0.03210 & \textbf{0.02791} & $[-0.00448,-0.00389]$\\
 & High motion & 0.03659 & \textbf{0.03288} & $[-0.00441,-0.00304]$\\
\midrule
\multirow{2}{*}{\shortstack[l]{Language\\Table}} & All & 0.03920 & \textbf{0.03580} & $[-0.00363,-0.00317]$\\
 & High motion & 0.04488 & \textbf{0.04165} & $[-0.00372,-0.00275]$\\
\bottomrule
\end{tabular}
\end{table}

%% file: tables/supp_table06_long_horizon.tex
\begin{table}[t]
\centering\footnotesize
\setlength{\tabcolsep}{3pt}
\caption{\textbf{33-frame rollouts.} TwoPass minus SC on Bridge and minus ordinary Wan on Language Table, all at 300k iterations and \nfe{4}. 128 Bridge and 512 Language Table cases; 95\% intervals.}
\label{tab:long_horizon}
\begin{tabular}{@{}lllrr@{}}
\toprule
Dataset & Window & Metric & Difference & 95\% interval\\
\midrule
\multirow{6}{*}{Bridge} & Second chunk & PSNR & +1.03207 & $[+0.74484,+1.32431]$\\
 & Second chunk & SSIM & +0.00434 & $[+0.00106,+0.00774]$\\
 & Second chunk & LPIPS & -0.00651 & $[-0.01005,-0.00283]$\\
\addlinespace[2pt]
 & Full future & PSNR & +0.83456 & $[+0.61340,+1.04821]$\\
 & Full future & SSIM & +0.00316 & $[+0.00055,+0.00554]$\\
 & Full future & LPIPS & -0.00463 & $[-0.00702,-0.00200]$\\
\midrule
\multirow{6}{*}{\shortstack[l]{Language\\Table}} & Second chunk & PSNR & +0.52192 & $[+0.37877,+0.65608]$\\
 & Second chunk & SSIM & +0.00334 & $[+0.00237,+0.00424]$\\
 & Second chunk & LPIPS & -0.00875 & $[-0.01066,-0.00663]$\\
\addlinespace[2pt]
 & Full future & PSNR & +0.52504 & $[+0.42565,+0.62489]$\\
 & Full future & SSIM & +0.00297 & $[+0.00229,+0.00364]$\\
 & Full future & LPIPS & -0.00713 & $[-0.00843,-0.00577]$\\
\bottomrule
\end{tabular}
\end{table}

%% file: tables/supp_table07_object_motion.tex
\begin{table}[t]
\centering\footnotesize
\setlength{\tabcolsep}{3.5pt}
\caption{\textbf{Object motion accuracy, \nfe{4}.} Errors in pixels (lower is better) and placement accuracy within 10 pixels (higher is better) of tracked moving objects, averaged over objects and then windows. Static keeps every object at its first-frame position. Bold marks the best model.}
\label{tab:object_motion}
\begin{tabular}{@{}llrrrr@{}}
\toprule
Subset & Method & Center$\downarrow$ & Displ.$\downarrow$ & Endpoint$\downarrow$ & Acc@10$\uparrow$\\
\midrule
\multirow{4}{*}{\shortstack[l]{Bridge,\\all moving}} & Static & 23.17 & 34.24 & 34.24 & 19.8\\
 & SC & 5.36 & 6.35 & 8.23 & 72.9\\
 & ORV & 5.18 & 5.69 & 7.88 & 78.7\\
 & \textbf{TwoPass} & \textbf{4.47} & \textbf{5.21} & \textbf{6.95} & \textbf{80.2}\\
\midrule
\multirow{3}{*}{\shortstack[l]{Bridge,\\manipulated}} & SC & 3.72 & 4.34 & 5.54 & 84.3\\
 & ORV & 3.91 & 4.50 & 6.01 & 84.9\\
 & \textbf{TwoPass} & \textbf{3.32} & \textbf{3.75} & \textbf{5.13} & \textbf{86.7}\\
\midrule
\multirow{3}{*}{\shortstack[l]{Bridge,\\arm}} & SC & 7.79 & 9.31 & 12.00 & 57.9\\
 & ORV & 7.19 & 7.71 & 10.69 & 68.6\\
 & \textbf{TwoPass} & \textbf{6.30} & \textbf{7.47} & \textbf{9.79} & \textbf{70.0}\\
\midrule
\multirow{3}{*}{\shortstack[l]{Language Table,\\all moving}} & Static & 37.64 & 51.94 & 51.94 & 15.1\\
 & Ordinary & 11.37 & 13.09 & 16.81 & 46.1\\
 & \textbf{TwoPass} & \textbf{10.10} & \textbf{12.02} & \textbf{15.37} & \textbf{52.4}\\
\bottomrule
\end{tabular}
\end{table}

%% file: tables/supp_table08_object_motion_diff.tex
\begin{table}[t]
\centering\footnotesize
\setlength{\tabcolsep}{6pt}
\caption{\textbf{Object motion errors, TwoPass minus the other method,} all moving objects, \nfe{4}, in pixels. Every difference is significant.}
\label{tab:object_motion_diff}
\begin{tabular}{@{}llrrr@{}}
\toprule
Dataset & Minus & Center & Displ. & Endpoint\\
\midrule
Bridge & SC & $-0.90$ & $-1.14$ & $-1.27$\\
Bridge & ORV & $-0.71$ & $-0.48$ & $-0.92$\\
Language Table & Ordinary & $-1.27$ & $-1.08$ & $-1.44$\\
\bottomrule
\end{tabular}
\end{table}

%% file: tables/supp_table09_action_discrimination.tex
\begin{table}[t]
\centering\footnotesize
\setlength{\tabcolsep}{8pt}
\caption{\textbf{Action discrimination on the complete test sets.} Percentage of windows in which the TwoPass prediction made with the recorded actions is closer to the ground-truth future than the prediction made with the alternative actions.}
\label{tab:action_discrimination}
\begin{tabular}{@{}llrr@{}}
\toprule
Dataset & Alternative & By PSNR & By LPIPS\\
\midrule
\multirow{2}{*}{Bridge} & Swapped & 99.63 & 99.59\\
 & Zeroed & 99.42 & 99.29\\
\midrule
\multirow{2}{*}{\shortstack[l]{Language\\Table}} & Swapped & 97.05 & 97.61\\
 & Zeroed & 96.36 & 96.59\\
\bottomrule
\end{tabular}
\end{table}

%% file: tables/supp_table10_action_ranking.tex
\begin{table}[t]
\centering\footnotesize
\setlength{\tabcolsep}{4pt}
\caption{\textbf{Candidate action ranking.} TwoPass at \nfe{4} ranks eight candidate action sequences per window by the distance of their prediction to the ground-truth future, on 300 windows per dataset. Top-1 is the percentage of windows in which the recorded actions rank first; MRR is the mean reciprocal rank.}
\label{tab:action_ranking}
\begin{tabular}{@{}llrrr@{}}
\toprule
Dataset & Distance & Top-1 (\%)$\uparrow$ & MRR$\uparrow$ & Mean rank$\downarrow$\\
\midrule
\multirow{3}{*}{Bridge} & LPIPS & 98.0 & 0.988 & 1.043\\
 & PSNR & 98.3 & 0.990 & 1.037\\
 & Changed-pixel MSE & 98.0 & 0.988 & 1.037\\
\midrule
\multirow{3}{*}{\shortstack[l]{Language\\Table}} & LPIPS & 94.7 & 0.969 & 1.087\\
 & PSNR & 93.3 & 0.956 & 1.173\\
 & Changed-pixel MSE & 93.7 & 0.961 & 1.130\\
\midrule
Chance & -- & 12.5 & 0.34 & 4.5\\
\bottomrule
\end{tabular}
\end{table}

%% file: tables/supp_table11_ranking_margin.tex
\begin{table}[t]
\centering\footnotesize
\setlength{\tabcolsep}{5pt}
\caption{\textbf{Margin of the recorded actions over the closest wrong candidate.} In dB for PSNR; for LPIPS and changed-pixel MSE, the margin divided by the distance of the recorded actions. Same 300 windows as Table~\ref{tab:action_ranking}.}
\label{tab:ranking_margin}
\begin{tabular}{@{}llrrr@{}}
\toprule
Dataset & Distance & \textbf{TwoPass} & Control & Gain\\
\midrule
\multirow{3}{*}{Bridge} & PSNR & \textbf{4.02} & 3.44 & $+0.58$\\
 & LPIPS & \textbf{0.92} & 0.81 & $+0.11$\\
 & Changed-pixel MSE & \textbf{1.97} & 1.57 & $+0.40$\\
\midrule
\multirow{3}{*}{\shortstack[l]{Language\\Table}} & PSNR & \textbf{2.62} & 2.06 & $+0.56$\\
 & LPIPS & \textbf{0.70} & 0.52 & $+0.18$\\
 & Changed-pixel MSE & \textbf{1.11} & 0.70 & $+0.41$\\
\bottomrule
\end{tabular}
\end{table}

%% file: tables/supp_table12_fine_ranking.tex
\begin{table}[t]
\centering\footnotesize
\setlength{\tabcolsep}{2.5pt}
\caption{\textbf{Fine-grained action ranking.} Eight candidates mix the recorded actions with another window's actions in proportions 0--100\%. Spearman is the rank correlation between mixing proportion and prediction distance; Pairwise is the percentage of the 28 candidate pairs in which the less-deviated candidate has the smaller distance. 300 windows per dataset.}
\label{tab:fine_ranking}
\begin{tabular}{@{}llcccc@{}}
\toprule
 & & \multicolumn{2}{c}{Spearman$\uparrow$} & \multicolumn{2}{c}{Pairwise (\%)$\uparrow$}\\
\cmidrule(lr){3-4}\cmidrule(l){5-6}
Dataset & Distance & \textbf{TwoPass} & Control & \textbf{TwoPass} & Control\\
\midrule
\multirow{3}{*}{Bridge} & LPIPS & \textbf{0.945} & 0.943 & \textbf{95.4} & 95.0\\
 & PSNR & \textbf{0.940} & 0.928 & \textbf{95.1} & 94.1\\
 & Changed-pixel MSE & \textbf{0.894} & 0.885 & \textbf{91.7} & 91.0\\
\midrule
\multirow{3}{*}{\shortstack[l]{Language\\Table}} & LPIPS & \textbf{0.855} & 0.816 & \textbf{89.8} & 87.3\\
 & PSNR & \textbf{0.811} & 0.789 & \textbf{86.9} & 85.3\\
 & Changed-pixel MSE & \textbf{0.834} & 0.803 & \textbf{88.0} & 86.1\\
\bottomrule
\end{tabular}
\end{table}

%% file: tables/supp_table13_mix_curve.tex
\begin{table*}[t]
\centering\footnotesize
\setlength{\tabcolsep}{5pt}
\caption{\textbf{Prediction error against action deviation.} PSNR and LPIPS of the prediction made with a candidate that mixes the recorded actions with another window's actions, by the proportion of the other actions; same 300 windows per dataset as Table~\ref{tab:fine_ranking}.}
\label{tab:mix_curve}
\begin{tabular}{@{}llrrrrrrrr@{}}
\toprule
Dataset, metric & Model & 0\% & 10\% & 20\% & 30\% & 50\% & 70\% & 85\% & 100\%\\
\midrule
\multirow{2}{*}{Bridge, PSNR$\uparrow$} & \textbf{TwoPass} & 25.20 & 24.35 & 23.15 & 22.15 & 20.82 & 20.05 & 19.71 & 19.48\\
 & SC & 24.54 & 23.84 & 22.84 & 21.96 & 20.72 & 19.98 & 19.65 & 19.43\\
\multirow{2}{*}{Bridge, LPIPS$\downarrow$} & \textbf{TwoPass} & 0.059 & 0.067 & 0.080 & 0.094 & 0.116 & 0.130 & 0.137 & 0.141\\
 & SC & 0.063 & 0.070 & 0.083 & 0.096 & 0.117 & 0.130 & 0.137 & 0.142\\
\midrule
\multirow{2}{*}{Language Table, PSNR$\uparrow$} & \textbf{TwoPass} & 26.61 & 25.67 & 24.71 & 24.03 & 23.16 & 22.67 & 22.46 & 22.30\\
 & Ordinary & 26.01 & 25.25 & 24.51 & 23.91 & 23.16 & 22.68 & 22.46 & 22.31\\
\multirow{2}{*}{Language Table, LPIPS$\downarrow$} & \textbf{TwoPass} & 0.056 & 0.063 & 0.074 & 0.084 & 0.098 & 0.107 & 0.112 & 0.116\\
 & Ordinary & 0.062 & 0.068 & 0.077 & 0.086 & 0.098 & 0.107 & 0.113 & 0.116\\
\bottomrule
\end{tabular}
\end{table*}

%% file: sections/module_studies.tex
\section{Component Differences at Shorter Training}
\label{sec:module_training}
The component chain of the main paper is repeated at shorter training budgets, 150k iterations on Bridge and 100k on Language Table, on the complete test sets at \nfe{4}. Self-conditioning raises PSNR over ordinary Wan by 0.13\,dB on Bridge and 0.18\,dB on Language Table, and the adapter adds 0.10\,dB and 0.13\,dB; SSIM and LPIPS improve along the chain on both datasets; FVD falls along the chain on Bridge and is level on Language Table, where the three variants lie within 0.15.

\section{Feedback-Path Controls}
\label{sec:role_studies}
Table~\ref{tab:core} compares SC with ordinary Wan, with a model whose draft pathway receives zeros throughout training (zero-input pathway), and with SC evaluated with a zero draft (input off), after 25k iterations. At \nfe{10}, SC improves PSNR, SSIM, LPIPS and FVD over ordinary training.
\input{tables/supp_table14_core}

\input{sections/training_depth_studies}

%% file: tables/supp_table14_core.tex
\begin{table}[t]
\centering\footnotesize
\setlength{\tabcolsep}{5pt}
\caption{\textbf{Feedback-path controls.} Bridge, 256 test cases, \nfe{10}, 25k iterations with 16-frame clips. Input off is an inference setting.}
\label{tab:core}
\begin{tabular}{@{}lrrrr@{}}
\toprule
Variant & PSNR$\uparrow$ & SSIM$\uparrow$ & LPIPS$\downarrow$ & FVD$\downarrow$\\
\midrule
Ordinary & 21.759 & 0.8313 & 0.10360 & 101.74\\
Zero-input pathway & 21.773 & 0.8291 & 0.10372 & 98.38\\
SC, input off & 21.794 & 0.8300 & 0.10338 & 97.64\\
SC & \textbf{22.076} & \textbf{0.8344} & \textbf{0.09894} & \textbf{93.27}\\
\bottomrule
\end{tabular}
\end{table}

%% file: sections/training_depth_studies.tex
\section{Training Pass Depth and Cost}
\label{sec:completed_npass}
Training depth $N$ is independent of the inference NFE. The depth study keeps noisy state, noise level, actions, and observed boundary fixed while applying $N-1$ detached draft passes before one supervised pass. The feedback probability is 0.5, giving $1+(N-1)/2$ expected network calls per update. Only the supervised pass builds an autograd graph. All evaluations here use \nfe{10} and 256 Bridge test cases. Training uses 16-frame clips. The depth comparison in the main paper continues training from a shared checkpoint; Tables~\ref{tab:core} and~\ref{tab:npass_official} start from a different one.

\paragraph{Matched iterations and matched time.}
At matched iterations in the main paper's depth table, three passes minus two change PSNR by $+0.02960$\,dB (95\% interval $[0.00991,0.05364]$) and four passes minus two by $+0.01452$\,dB ($[-0.01405,0.04247]$).

\paragraph{Depth across training lengths.}
Table~\ref{tab:npass_official} trains each depth for 50k to 100k iterations. At 50k and at 100k, two passes reach the highest PSNR and SSIM and the lowest LPIPS, and at every training length they take the least training time, which supports $N=2$.
\input{tables/supp_table15_npass_official}

%% file: tables/supp_table15_npass_official.tex
\begin{table}[t]
\centering\footnotesize
\setlength{\tabcolsep}{5pt}
\caption{\textbf{Training depth at three training lengths.} Bridge, 256 test cases, \nfe{10}; 16-frame training clips. Time (h) is H100 GPU-hours divided by eight.}
\label{tab:npass_official}
\begin{tabular}{@{}rlrrrr@{}}
\toprule
Iter. & $N$ & PSNR$\uparrow$ & SSIM$\uparrow$ & LPIPS$\downarrow$ & Time (h)$\downarrow$\\
\midrule
50k & 2 & \textbf{23.193} & \textbf{0.8478} & \textbf{0.08428} & \textbf{5.169}\\
50k & 3 & 22.981 & 0.8472 & 0.08624 & 5.715\\
50k & 4 & 23.163 & \textbf{0.8478} & 0.08457 & 6.210\\
\addlinespace[2pt]
75k & 2 & 23.578 & \textbf{0.8531} & 0.07950 & \textbf{7.784}\\
75k & 3 & \textbf{23.619} & 0.8528 & \textbf{0.07922} & 8.631\\
75k & 4 & 23.592 & 0.8527 & 0.07929 & 9.422\\
\addlinespace[2pt]
100k & 2 & \textbf{23.864} & \textbf{0.8557} & \textbf{0.07638} & \textbf{10.447}\\
100k & 3 & 23.818 & 0.8545 & 0.07683 & 11.573\\
\bottomrule
\end{tabular}
\end{table}